\documentclass[10pt,twocolumn,letterpaper]{article}

\usepackage{cvpr}
\usepackage{times}
\usepackage{epsfig}
\usepackage{graphicx}
\usepackage{subfigure}
\usepackage{amsmath}
\usepackage{amssymb}
\usepackage{cases}

\usepackage{times}
\usepackage{epsfig}
\usepackage{graphicx}
\usepackage{amsmath}
\usepackage{amssymb}
\usepackage[linesnumbered,ruled,vlined,lined,boxed,commentsnumbered]{algorithm2e}
\usepackage{subfigure}
\usepackage{indentfirst}
\usepackage{multirow}
\usepackage{booktabs}
\usepackage{threeparttable}
\usepackage{boldline}
\usepackage{float}
\usepackage{tabularx}
\usepackage{makecell}

\newcommand{\udots}{\mathinner{\mskip1mu\raise1pt\vbox{\kern7pt\hbox{.}}\mskip2mu\raise4pt\hbox{.}\mskip2mu\raise7pt\hbox{.}\mskip1mu}}

\usepackage[pagebackref=true,breaklinks=true,letterpaper=true,colorlinks,bookmarks=false]{hyperref}

\cvprfinalcopy % *** Uncomment this line for the final submission

\def\cvprPaperID{3450} % *** Enter the CVPR Paper ID here

\begin{document}

%%%%%%%% TITLE
\title{Solution for Large-Scale Hierarchical Object Detection Datasets with Incomplete Annotation and Data Imbalance}

\author{Yuan Gao, Xingyuan Bu, Yang Hu, Hui Shen, Ti Bai,  Xubin Li and Shilei Wen \\
VideoPlus Team at Department of Computer Vision Technology, Baidu \\
corresponding author: wenshilei@baidu.com}

\maketitle
%\thispagestyle{empty}

%%%%%%%%% ABSTRACT
\begin{abstract}
This report demonstrates our solution for the Open Images 2018 Challenge. Based on our detailed analysis on the Open Images Datasets (OID), it is found that there are four typical features: \textbf{large-scale, hierarchical tag system, severe annotation incompleteness and data imbalance}. Considering these characteristics, an amount of strategies are employed, including SNIPER, soft sampling, class-aware sampling (CAS), hierarchical non-maximum-suppression (HNMS) and so on. In virtue of these effective strategies, and further using the powerful SENet154 armed with feature pyramid module and deformable ROI-align as the backbone, our best single model could achieve a mAP of 56.9\%. After a further ensemble with 9 models, the final mAP is boosted to 62.2\% in the public leaderboard (ranked the 2nd place) and 58.6\% in the private leaderboard (ranked the 3rd place, slightly inferior to the 1st place by only 0.04 point).
\end{abstract}

%%%%%%%%% BODY TEXT
%%%%%%%%% INTRODUCTION
\section{Introduction}
\label{introduction}
To better understand the visual content, we should not only know \textit{what} is the object, i.e, the so-called classification task, but also know \textit{where} is the very object, i.e., the so-called location task. The object detection task is to simultaneously provide these two information for a given image. 

Depending on the pipeline, most of the object detection techniques could be divided into two categories, i.e., one-stage method \cite{redmon2016you,lin2018focal,redmon2017yolo9000} and two-stage method \cite{girshick2014rich,he2014spatial,girshick2015fast,ren2015faster,he2017mask,dai2016r,li2017light,li2018detnet,liu2018path,cai2017cascade,jiang2018acquisition,hu2017relation}. Generally speaking, the one-stage methods focus on the  detection speed while the dominant merit of the two-stage methods is the detection precision. In this challenge, we concentrate on the two-stage methods considering the advantages in detection precision. 

Specifically, in the modern convolutional neural network (CNN) context, the regions with CNN features (R-CNN) \cite{girshick2014rich} method should be the earliest two-stage detector. Just as its name implies, the R-CNN methods first output multiple region proposals using the selective-search algorithm, then regress the bounding-box (bbox) coordinates and classify each proposal into a specified class based on the extracted CNN features using the matured support vector machine (SVM) algorithm. To accelerate the pipeline, the SPPNet \cite{he2014spatial} is proposed by claiming that the feature maps could be shared by different proposals, and hence reducing the computation burden of the feature extraction process. Similar idea is used by the well-known Fast R-CNN \cite{girshick2015fast} method. In this method, the features of the proposed regions are extracted by a newly-designed region-of-interest pooling (ROI-pooing) layer, and a multitask loss which combines the regression loss and the classification loss is considered for optimized training process. It should be noted that for all the above mentioned methods, the regions are proposed in an offline method such that they could not be end-to-end optimized in one network. To solve this problem and therefore enable an end-to-end training style, a region proposal network (RPN) is incorporated into the overall pipeline, shaping the well-known Faster R-CNN method \cite{ren2015faster}. It should be noted that the RPN is nearly cost-free considering the backbone-sharing property. Aforementioned improvements of two-stage detection algorithms mainly focus on speeding up the whole pipeline. 

Another track of improving focuses on boosting precision of detectors \cite{li2017light,li2018detnet,liu2018path,cai2017cascade,jiang2018acquisition,hu2017relation}. As we know, the series of R-CNN based methods uses the same feature maps to handle both the large and small objects, and consequently cannot adapt the object scales. To alleviate this drawback, one can either use the image-level or the feature-level solutions. As for the image-level solution, an intuitive method is to use the multiscale training/testing (MST) strategy. However, as pointed out by Bharat Singh \textit{et.al.}, the MST strategy are trying to memorize the features of objects with different scales based on the capacity of the network, resulting in a capacity waste. Considering this observation, they proposed a nice solution, called SNIPER \cite{singh2018sniper}, by feeding the network the samples with similar scale, and hence could make better use of the capacity. As for the feature-level solution, the feature pyramid network (FPN) \cite{lin2017feature} is proposed to construct multiscale features with rich semantic information by designing a top-down architecture, and has been a standard module in modern CNN-based detectors. On the other hand, to further use the available segmentation mask information, except for the classification and regression heads in the Faster R-CNN framework, an extra mask head is added in the Mask R-CNN \cite{he2017mask} method which results in the state-of-the-art algorithm performance.

The detection algorithms are pushing forward to be faster and more precise by talented researchers. However, the bounding box annotation in the detection task is much more expensive compared to the label annotation in the classification task. As a result, the dataset scale for the detection task is still relatively small compared to that for the classification task, thus limiting the performance of the detection task. To alleviate this problem, Google has open-sourced the Open Images Dataset (OID) \cite{openimages} in the Open Images challenge. 
\begin{figure*}
\centering
\includegraphics[width=0.7\textwidth]{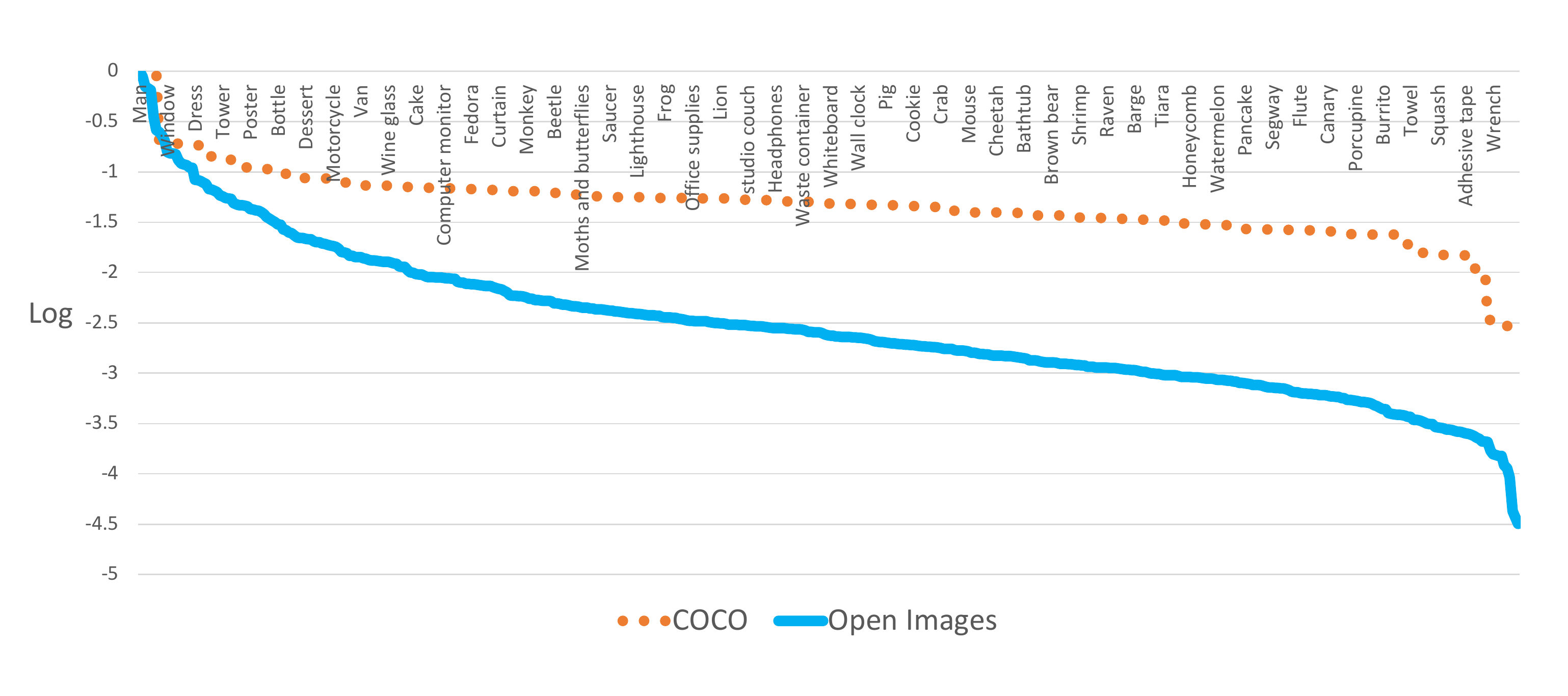}
\label{fig:databalance}
\caption{The data statistics of the Open Images and the MS-COCO datasets. The x-axis and the y-aixs are with the label and the log-transformed instance counts, respectively. It should be noted that the label number of the Open Images and the MS-COCO datasets are different, which is 500 and 80, respectively. For better visualization, we have duplicate the statistics of the MS-COCO datasets by a mean value of 6.25 (some are duplicated by 6 times, some are 7).}
\end{figure*}

\begin{figure*}
\centering
\includegraphics[width=0.7\textwidth]{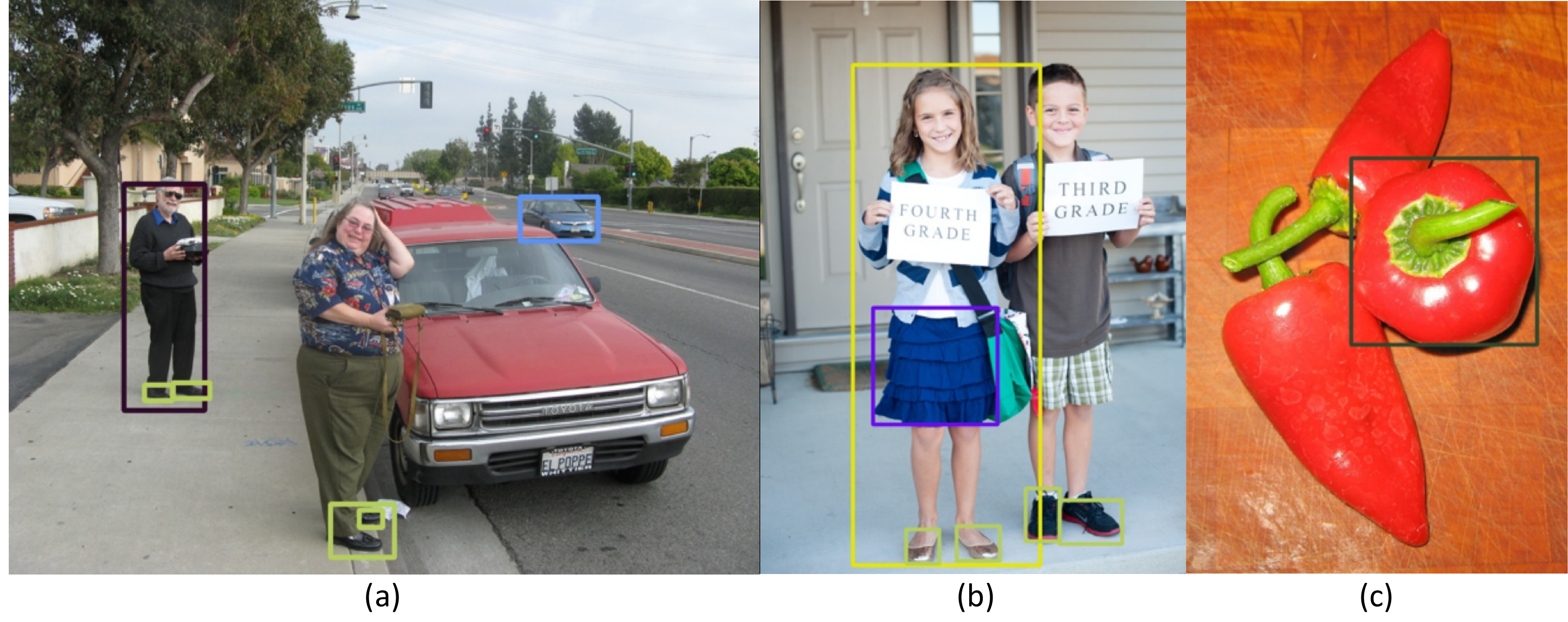}
\label{fig:label_missing}
\caption{Annotation incompleteness visualization. In (a), the missed annotation is the person and the car. In (b), this missed annotation is the person. In (c), the missed annotation is the pepper.}
\end{figure*}
Based on our detailed analysis, it is found that there exist several characteristics about this dataset: \\
\textbf{Large-scale} The OID contains 1.7 million images with bounding box annotations, containing 12 million instances. In contrast, the well-known COCO dataset contain around 110K images.\\
\textbf{Hierarchical tag system} This dataset contains 500 categories, consisting of 5 different levels, where the first level has 177 categories. \\
\textbf{Annotation incompleteness} As this dataset is too large to be annotated completely, there exist severe missed annotations for the bounding box, as demonstrated in figure 2. \\
\textbf{Data imbalance} As shown in figure~\ref{fig:databalance}, compared to the COCO dataset, the OID exhibits much more severe data imbalance. For example, the category of \textit{person} has 1.4 million instances, which is $10^{5}$ larger than that of the \textit{pressure cooker} which only has 14 instances.

Regarding these characteristics, in our solution, an amount of effective strategies are adopted, such as SNIPER, soft sampling, class-aware sampling (CAS), hierarchical non-maximum-suppression (HNMS) and so on. Leveraging these strategies, we achieve a 56.9\% mAP for our best single model. After ensemble with 9 different models, our final mAP is boosted to 62.2\% in the public leaderboard and 58.6\% in the private leaderboard.

\section{Method}
\label{method}
\begin{figure*}
\centering
\includegraphics[width=0.9\textwidth]{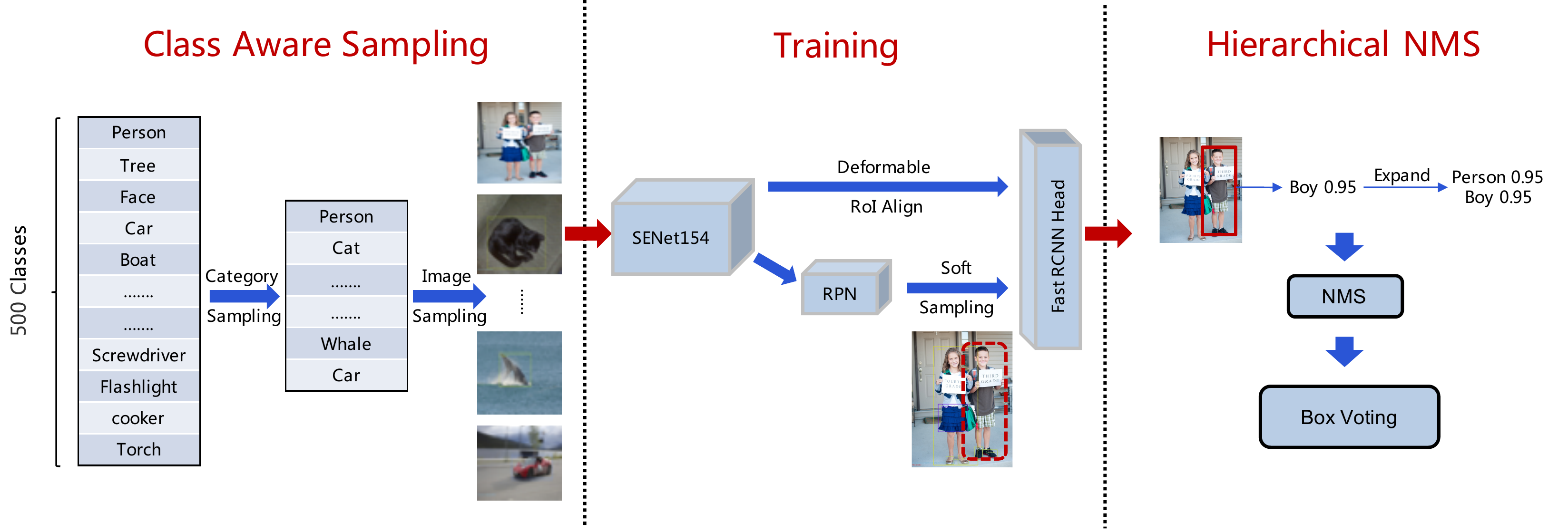}
\caption{The workflow of our detection system.}
\label{fig:workflow}
\end{figure*}
In this section, we will detailedly demonstrate our strategies used in this challenge. The overall workflow could be find in figure~\ref{fig:workflow}. And also, we would also like to remark the tricks which are proved to be useful in many datasets but failed in this challenge. 

\subsection{Baseline model}
\label{baseline}
In this challenge, the Faster R-CNN framework is adopted as the baseline, where the backbone is ResNet101. To handle the objects in different scales, the FPN module is added. With this basic configuration, we achieve a 43\% mAP as our baseline. 

\subsection{Deformable ROI-Align}
As mentioned in section \ref{introduction}, there are 500 categories in this dataset. And the shapes of the objects are also colorful. To handle this complexity more effectively, the powerful deformable ROI-align \cite{dai2017deformable} is utilized to enhance the performance of our baseline model. This layer could improve the model by 0.5 point to 43.5\%. 

\subsection{SNIPPER}
The used large-scale dataset contains 1.7 million images, which exhibits a main roadblock for the model interactive. To enable a fast trial-and-error idea check, we use the SNIPER strategy to accelerate our model training. Basically, the SNIPER method is an image-pyramid training strategy by cropping a certain region from the origin image based on the annotated groundtruth bounding boxes, and then feeding this \textit{small} region into the network for training. Considering the relative small input size, we can use larger batch size. Consequently, the batch normalization (BN) could be used, and also a higher GPU efficiency is achieved. This could result in much faster training speed. Basically, the training process will be converged in 12 hours by using 8 NVIDIA Tesla P40 GPUs. On the other hand, as the selected region is selected based on the groundtruth annotations, the SNIPPER method also could be useful to handle the annotation-incompleteness problems. 

\subsection{Soft sampling}
The annotation incompleteness problem is another degradation factor in this dataset, as shown in figure 2. Consequently, 	a potential positive region indeed containing the object but not being annotated will be regarded as the negative sample. This conflict will pose severe difficulties to our model training. To solve this problem, the soft sampling strategy \cite{zhe2018soft} is used by reweighting the negative samples based on the intersection-over-union (IOU) with the groundtruth boxes. 

\subsection{Class aware sampling}
The data statistics of the dataset is demonstrated in figure~\ref{fig:databalance}, which exhibits much severe data imbalance compared to the COCO dataset. In this challenge, a class aware sampling method \cite{shen2016relay} is adopted to alleviate this problem. In details, as shown in figure~\ref{fig:workflow}, the class aware sampling is consisted of two stages: category level sampling and image level sampling. Firstly, a number of categories as the batch size is uniformed sampled from the total 500 categories. And then, for each category, we sample one image from the images containing objects with the specified class. This strategy could make sure all the category has the same chance to be sampled, and hence the model could be trained sufficiently. With this strategy, the performance of our detector is improved greatly, showing a 7.6 point mAP increase.

\subsection{Hierarchical NMS}
\label{HNMS}
In the Open Images challenges, if a label in the child node is assigned to an instance, it implies that all the labels in the ancestor nodes are also assigned. Consequently, during the evaluation, the model should recall all the labels in its ancestor. However, due to the missed labels in the training datasets and the intra-class competition, one could not output all the labels in the child and ancestor nodes for the same bounding box. To solve this problem, a hierarchical NMS strategy is proposed. In details, given all the bounding boxes of an image predicted by the model, we expand the associated single label of a single bounding box to multi-labels with all the corresponded labels in the ancestor nodes, where the score is same as that of the child node. Based on these expanded results, a classical NMS pipeline is applied, where the threshold of the IOU is set as 0.5. On the other hand, if the IOU of two bounding boxes belonging to the same category is larger enough, say 0.9, one should have more confidence regarding the existence of the object in this location. As a consequence, the bounding box with the highest score should be increased further based on the score of the dropped bounding box. In this challenge, a 30\% score is voting to the bounding box with higher confidence. With the HNMS strategy, a 2 point mAP improvement is achieved.

\subsection{Other tricks}
\noindent\textbf{Training dataset augmentation} The dataset of the Open Images 2018 challenge contains 1.7 million images ranging in 500 categories, where 100K images are the official suggested validation dataset. In our custom settings, to accelerate the evaluation process and also enlarge the training dataset, we only use 5000 images as the mini-validation dataset, and the rest is used to augment the official training set. This simple adjustment could provide a 0.9 point improvement. \\
\textbf{More powerful backbone} To further strengthen the feature, the SENet 154 is adopted, improving the mAP by 1.8 point. \\
\textbf{MST} A popular strategy for detector training is using the multiscale training and testing. In this challenge, this strategy is also employed, resulting a 1.1 point improvement. \\
\textbf{Ensemble} For the final results, we have used 9 models for ensemble: one of them is ResNet101-FPN with SNIPER, and alternating the backbone between SENet154 and ResNeXt152 as well as switching on/off the deformable ROI-align and CAS components result in the rest 8 models. These could improve the performance from our best single model with 56.9\% to the final result with 62.2\% in the public leaderboard. \\
\textbf{OHEM} The online hard-example mining is a very popular strategy commonly used in the detection tasks. However, in this challenge, we find strong negative effect on the final results. Based on our analysis, the main reason should be that the severe label missing in the annotations of the training datasets. Qualitatively, if one instance is correctly detected by our model, but it is not annotated as groundtruth, such instance will be falsely treated as hard negative example in OHEM and the optimizer will push it into the wrong direction, resulting in degraded performance. 

\section{Results}
\label{result}
During the training process, we just follow the typical hyper-parameter settings. Generally speaking, the initial learning rate is 0.06 and reduced to 0.006 after 40K iterations. The whole training process will be terminated after 60K iterations. 48 Tesla-V100 GPU are utilized for training, and the batch size of 48 is used during training.

\begin{table}
\caption{Performance improvement by adding different strategies step by step.}
%\hline %\vspace{0.08cm}
\begin{tabular}{l|c}
\vspace{0.08cm}
\textbf{Method} & \textbf{Public Leader board}\\
\toprule
ResNet101 Faster-R-CNN FPN & 43 \\
+Deformable RoI Align & 43.5 \\
+Validation Data & 44.4 \\
+SENet154 & 46.2 \\
+Multi-Scale Training and Testing & 47.3 \\
+Class Aware Sampling(CAS) & 54.9 \\
+Hierarchical NMS(HNMS) & 56.9 \\
+Ensemble (SNIPER, w/o CAS, ...) & 62.2 \\
\hline
\end{tabular}
\end{table}

Table 1 demonstrates the detailed improvements with different training strategies. As can be seen, the CAS strategy could boost the performance heavily, i.e., 7.6 points. By further using the HNMS strategy, the performance could be further improved by 2 points, achieving the best single best model with a mAP of 56.9\%. With a final ensemble strategy with 9 different models, we achieve the 62.2\% mAP performance in the public leaderboard, ranking the 2nd place. 

\begin{figure*}
\centering
\includegraphics[width=0.9\textwidth]{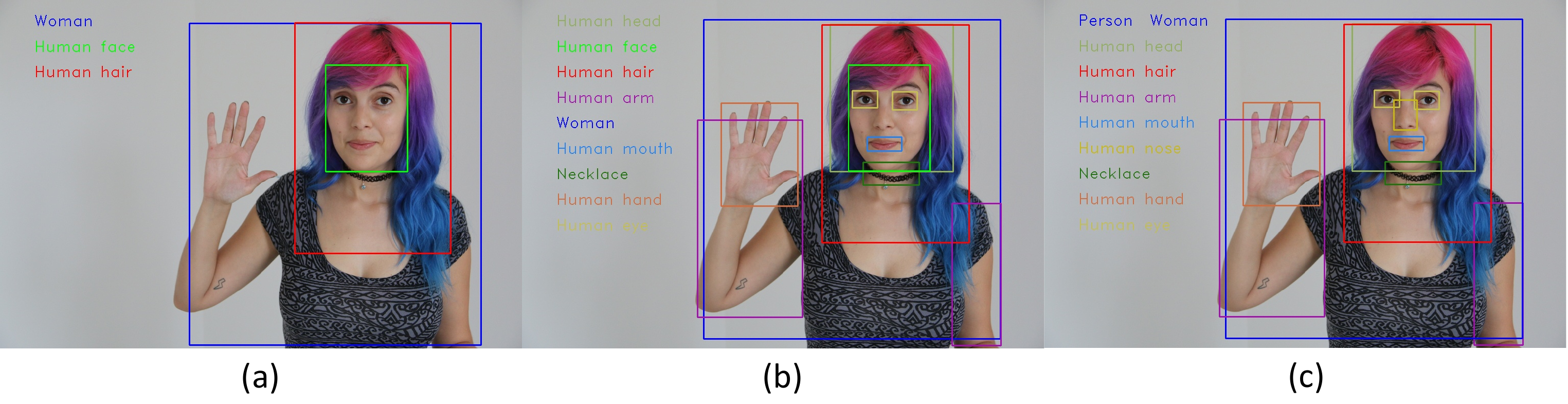}
\caption{Visualization of our single model with different strategies. Figures (a) to (c) correspond to the baseline model, baseline model with CAS strategy, baseline model with the CAS and HNMS strategies. The words in the figures are the associated labels detected. It should be noted that the label in the ancestor-child node relationship shares the same color.}
\label{fig:image_vis}
\end{figure*}
Figure~\ref{fig:image_vis} illustrates a visual comparison among different single models, i.e., baseline, baseline+CAS, baseline+CAS+HNMS. For the majority categories, such as the \textit{Human face} and \textit{Human hair}, all the models show good results. However, for the minor categories, such as the \textit{Human nose, Human eye, Human mouth} and so on, the baseline model cannot produce correct results. On the contrast, the CAS strategy could greatly alleviate these problems as demonstrated in figure~\ref{fig:image_vis} (b). With further HNMS strategy, the label from the ancestor nodes could be correctly predicted,  as shown that both the \textit{Person} and the \textit{Woman} are given in figure~\ref{fig:image_vis} (c).

\section{Conclusion}
\label{conclusion}
In this Open Image challenge, the dataset is in large scale with a hierarchical tag system. It is found that there exists severe data imbalance and annotation incompleteness problem. Taking these characteristics into consideration, a amount of strategies are employed, resulting a 56.9\% mAP for our best single model. After ensemble, the mAP is boosted to 62.2\%.  

%% We do not need to have new page for this.
%\newpage
{\small
\bibliographystyle{ieee}
\bibliography{intro_final}

\begin{thebibliography}{10}\itemsep=-1pt

\bibitem{cai2017cascade}
Z.~Cai and N.~Vasconcelos.
\newblock Cascade r-cnn: Delving into high quality object detection.
\newblock {\em arXiv preprint arXiv:1712.00726}, 2017.

\bibitem{dai2016r}
J.~Dai, Y.~Li, K.~He, and J.~Sun.
\newblock R-fcn: Object detection via region-based fully convolutional
  networks.
\newblock In {\em Advances in neural information processing systems}, pages
  379--387, 2016.

\bibitem{dai2017deformable}
J.~Dai, H.~Qi, Y.~Xiong, Y.~Li, G.~Zhang, H.~Hu, and Y.~Wei.
\newblock Deformable convolutional networks.
\newblock In {\em Proceedings of the IEEE International Conference on Computer
  Vision}, pages 764--773, 2017.

\bibitem{girshick2015fast}
R.~Girshick.
\newblock Fast r-cnn.
\newblock In {\em Proceedings of the IEEE international conference on computer
  vision}, pages 1440--1448, 2015.

\bibitem{girshick2014rich}
R.~Girshick, J.~Donahue, T.~Darrell, and J.~Malik.
\newblock Rich feature hierarchies for accurate object detection and semantic
  segmentation.
\newblock In {\em Proceedings of the IEEE conference on computer vision and
  pattern recognition}, pages 580--587, 2014.

\bibitem{he2017mask}
K.~He, G.~Gkioxari, P.~Doll{\'a}r, and R.~Girshick.
\newblock Mask r-cnn.
\newblock In {\em Computer Vision (ICCV), 2017 IEEE International Conference
  on}, pages 2980--2988. IEEE, 2017.

\bibitem{he2014spatial}
K.~He, X.~Zhang, S.~Ren, and J.~Sun.
\newblock Spatial pyramid pooling in deep convolutional networks for visual
  recognition.
\newblock In {\em European conference on computer vision}, pages 346--361.
  Springer, 2014.

\bibitem{hu2017relation}
H.~Hu, J.~Gu, Z.~Zhang, J.~Dai, and Y.~Wei.
\newblock Relation networks for object detection.
\newblock {\em arXiv preprint arXiv:1711.11575}, 8, 2017.

\bibitem{jiang2018acquisition}
B.~Jiang, R.~Luo, J.~Mao, T.~Xiao, and Y.~Jiang.
\newblock Acquisition of localization confidence for accurate object detection.
\newblock {\em arXiv preprint arXiv:1807.11590}, 2018.

\bibitem{openimages}
I.~Krasin, T.~Duerig, N.~Alldrin, V.~Ferrari, S.~Abu-El-Haija, A.~Kuznetsova,
  H.~Rom, J.~Uijlings, S.~Popov, S.~Kamali, M.~Malloci, J.~Pont-Tuset, A.~Veit,
  S.~Belongie, V.~Gomes, A.~Gupta, C.~Sun, G.~Chechik, D.~Cai, Z.~Feng,
  D.~Narayanan, and K.~Murphy.
\newblock Openimages: A public dataset for large-scale multi-label and
  multi-class image classification.
\newblock {\em Dataset available from
  https://storage.googleapis.com/openimages/web/index.html}, 2017.

\bibitem{li2017light}
Z.~Li, C.~Peng, G.~Yu, X.~Zhang, Y.~Deng, and J.~Sun.
\newblock Light-head r-cnn: In defense of two-stage object detector.
\newblock {\em arXiv preprint arXiv:1711.07264}, 2017.

\bibitem{li2018detnet}
Z.~Li, C.~Peng, G.~Yu, X.~Zhang, Y.~Deng, and J.~Sun.
\newblock Detnet: A backbone network for object detection.
\newblock {\em arXiv preprint arXiv:1804.06215}, 2018.

\bibitem{lin2017feature}
T.-Y. Lin, P.~Doll{\'a}r, R.~B. Girshick, K.~He, B.~Hariharan, and S.~J.
  Belongie.
\newblock Feature pyramid networks for object detection.
\newblock In {\em CVPR}, volume~1, page~3, 2017.

\bibitem{lin2018focal}
T.-Y. Lin, P.~Goyal, R.~Girshick, K.~He, and P.~Doll{\'a}r.
\newblock Focal loss for dense object detection.
\newblock {\em IEEE transactions on pattern analysis and machine intelligence},
  2018.

\bibitem{liu2018path}
S.~Liu, L.~Qi, H.~Qin, J.~Shi, and J.~Jia.
\newblock Path aggregation network for instance segmentation.
\newblock In {\em Proceedings of the IEEE Conference on Computer Vision and
  Pattern Recognition}, pages 8759--8768, 2018.

\bibitem{redmon2016you}
J.~Redmon, S.~Divvala, R.~Girshick, and A.~Farhadi.
\newblock You only look once: Unified, real-time object detection.
\newblock In {\em Proceedings of the IEEE conference on computer vision and
  pattern recognition}, pages 779--788, 2016.

\bibitem{redmon2017yolo9000}
J.~Redmon and A.~Farhadi.
\newblock Yolo9000: better, faster, stronger.
\newblock {\em arXiv preprint}, 2017.

\bibitem{ren2015faster}
S.~Ren, K.~He, R.~Girshick, and J.~Sun.
\newblock Faster r-cnn: Towards real-time object detection with region proposal
  networks.
\newblock In {\em Advances in neural information processing systems}, pages
  91--99, 2015.

\bibitem{shen2016relay}
L.~Shen, Z.~Lin, and Q.~Huang.
\newblock Relay backpropagation for effective learning of deep convolutional
  neural networks.
\newblock In {\em European conference on computer vision}, pages 467--482.
  Springer, 2016.

\bibitem{singh2018sniper}
B.~Singh, M.~Najibi, and L.~S. Davis.
\newblock Sniper: Efficient multi-scale training.
\newblock {\em arXiv preprint arXiv:1805.09300}, 2018.

\bibitem{zhe2018soft}
Z.~e.~a. Wu.
\newblock Soft sampling for robust object detection.
\newblock {\em arXiv preprint arXiv:1806.06986}, 2018.

\end{thebibliography}
}

\end{document}